\documentclass[letterpaper, 10 pt, conference]{ieeeconf}  

\IEEEoverridecommandlockouts                              

\usepackage{graphics}
\usepackage{epsfig} 
\usepackage{amsmath} 
\usepackage{amssymb}  
\usepackage{graphicx}
\usepackage[numbers]{natbib}
\usepackage{array}
\newcolumntype{M}[1]{>{\centering\arraybackslash}m{#1}}

\title{\LARGE \bf
Random Forests versus Neural Networks $-$ \linebreak What's Best for Camera Localization?
}

\author{\linespread{1.1}{\centering
Daniela Massiceti$^{1}$, Alexander Krull$^{2}$, Eric Brachmann$^{2}$, Carsten Rother$^{2}$ and Philip H.S. Torr$^{1}$}\\ 
\thanks{$^{1}$Faculty of Engineering Science, University of Oxford, United Kingdom
        {\tt\footnotesize \{daniela.massiceti, philip.torr\}@eng.ox.ac.uk}}%
\thanks{$^{2}$Fakult\"{a}t Informatik, Technische Universit\"{a}t Dresden, Germany
        {\tt\footnotesize \{alexander.krull, eric.brachmann, carsten.rother\}\newline @tu-dresden.de}}%
}

\begin{document}

\maketitle
\thispagestyle{empty}
\pagestyle{empty}

\begin{abstract}
This work addresses the task of camera localization in a known 3D scene given a single input RGB image. State-of-the-art approaches accomplish this in two steps: firstly, regressing for every pixel in the image its 3D scene coordinate and subsequently, using these coordinates to estimate the final 6D camera pose via RANSAC. To solve the first step, Random Forests (RFs) are typically used. On the other hand, Neural Networks (NNs) reign in many dense regression tasks, but are not test-time efficient. We ask the question: which of the two is best for camera localization? To address this, we make two method contributions: (1) a test-time efficient NN architecture which we term a ForestNet that is derived and initialized from a RF, and (2) a new fully-differentiable robust averaging technique for regression ensembles which can be trained end-to-end with a NN. Our experimental findings show that for scene coordinate regression, traditional NN architectures are superior to test-time efficient RFs and ForestNets, however, this does not translate to final 6D camera pose accuracy where RFs and ForestNets perform slightly better. To summarize, our best method, a ForestNet with a robust average, which has an equivalent fast and lightweight RF, improves over the state-of-the-art for camera localization on the 7-Scenes dataset \cite{Shotton2013}. While this work focuses on scene coordinate regression for camera localization, our innovations may also be applied to other continuous regression tasks.
\end{abstract}


\section{INTRODUCTION}

\begin{figure}[thpb]
\centering
\includegraphics[scale=0.3]{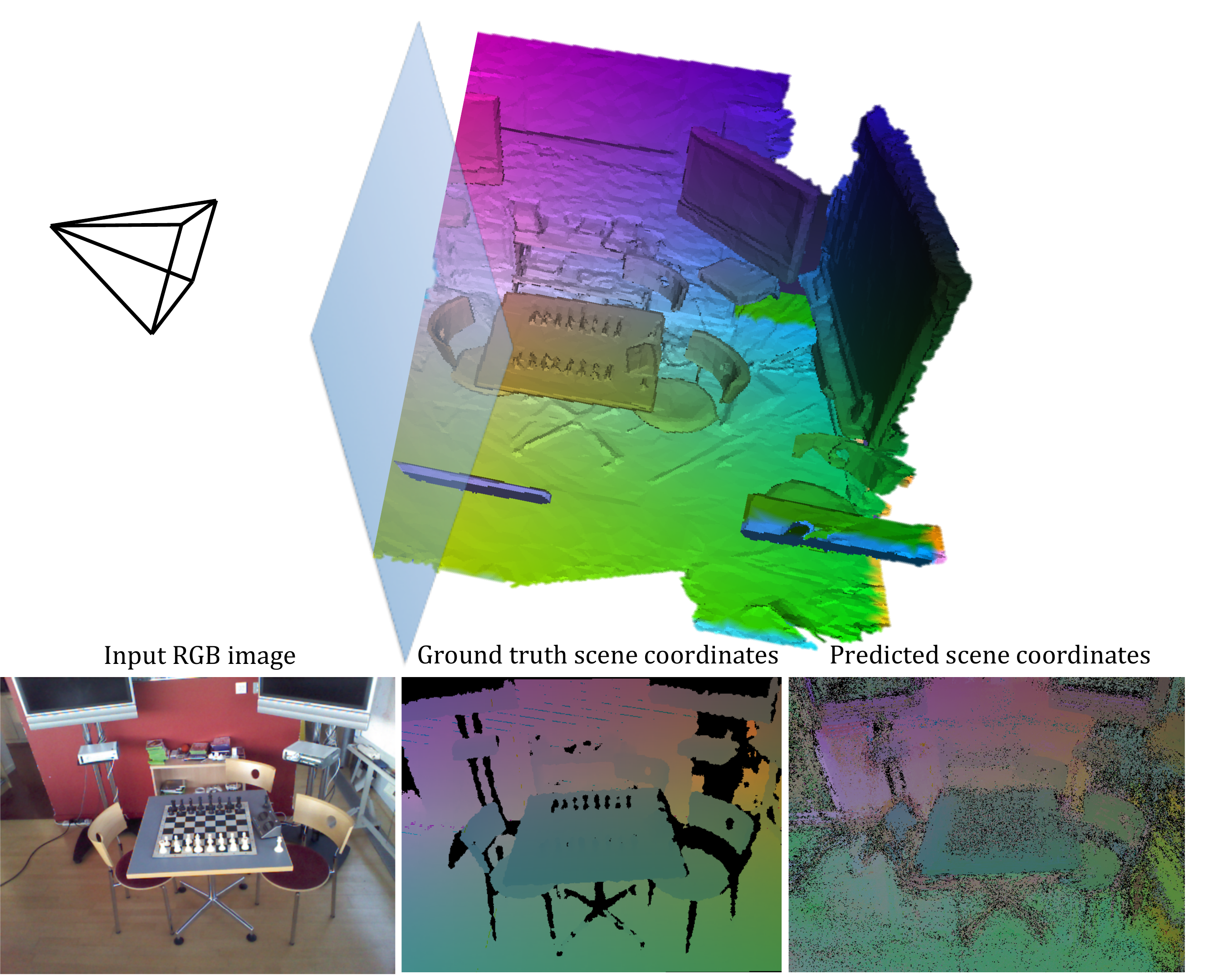}
\caption{\textbf{The concept of scene coordinate regression.} (Top) Given a known scene where each surface point has a unique 3D coordinate (here shown as a distinctive color), the goal is to locate the camera with respect to this scene. (Bottom) To achieve this, we need to predict at test time the scene coordinate for each pixel in an input RGB image (black pixels in the ground truth correspond to unknown scene coordinates).}
\label{fig:coolimage}
\end{figure}

\begin{figure*}[thpb]
\centering
\vskip 2mm
\includegraphics[scale=0.6]{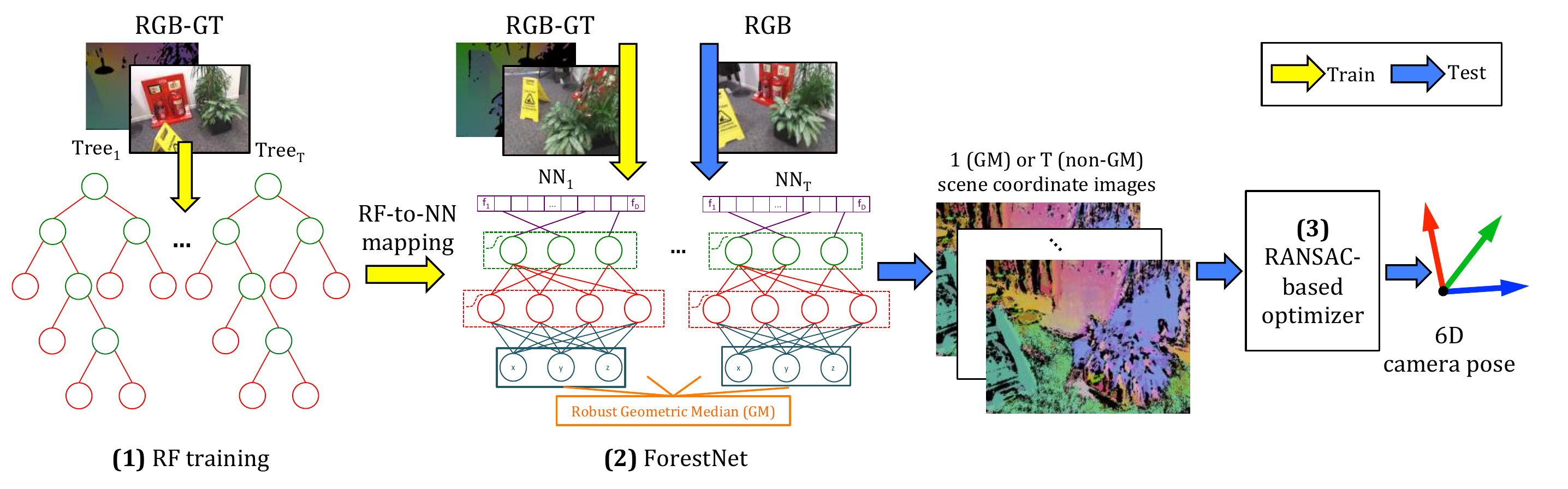}
\caption{\textbf{Training (yellow) and test (blue) pipeline using a ForestNet for camera localization}. (1) A RF is trained for a known 3D scene. (2) A RF-to-NN mapping is used to transform the trained RF into an ensemble of parallel tree-like networks, collectively referred to as a ForestNet. A subset of the original training data is used to fine-tune the ForestNet, with an additional robust averaging module which can be appended to and integrated into the training of the network. At test time, a sparse set of pixels is sampled from a RGB test image and passed through the ForestNet to obtain a set of 3D scene coordinates (either robustly averaged or not). (3) The scene coordinates are passed to the RANSAC optimizer of \cite{Brachmann2016} which generates and then refines a set of camera pose hypotheses until a final 6D camera pose remains. GT = Ground Truth Scene Coordinates. Best viewed in color.}
\label{fig:pipeline}
\end{figure*}

Given an RGB image of a known 3D scene, our goal is to infer the 6 degree of freedom pose of the camera - a task known as camera localization. To do this, current state-of-the-art methods \cite{Brachmann2016}, employ a two-stage pipeline. In the first stage, for each pixel in the image, a continuous 3D coordinate of the scene is regressed. These are called scene coordinates and uniquely identify each part of the scene (Fig.~\ref{fig:coolimage}). In the second stage, sparse sets of four scene coordinates are sampled and fed into a RANSAC-based optimizer which estimates the final camera pose.
In this work, we analyze the first stage of this pipeline, the dense regression of scene coordinates. State-of-the-art methods have typically employed regression Random Forests (RFs) to accomplish this task \cite{Shotton2013,Brachmann2016, Valentin2015}. The RFs have the advantage of being both memory and test-time efficient, two important features when considering camera localization in real-time navigation and mobile robotics. In this work, we analyze equivalently-efficient and non-efficient neural network (NN)-based architectures to accomplish the task of scene coordinate regression for camera localization. Our choices of architectures are inspired by the work of \cite{Sethi1990} who propose a method for transforming a binary decision tree into a two-hidden-layer NN. We utilise this tree-to-NN mapping by taking a RF trained for 2D-to-3D scene coordinate prediction in a known scene \cite{Brachmann2016} and mapping each of its trees to an equivalent two-layer NN. We collectively refer to this ensemble of NNs as a ForestNet. We fine-tune the ForestNet on a subset of the data that was used to train the original RF. At test-time, we obtain scene coordinate predictions from the ForestNet and feed them into a RANSAC-based optimizer to extract the 6D camera pose. This pipeline is illustrated in Fig.~\ref{fig:pipeline}.

The motivation for exploring NN architectures based on this RF-to-NN mapping is two-fold. Firstly, the mapping constructs a NN that is derived from a learned RF: the tree topologies define the active connections in the network and the trees' learned parameters specify the initialization of all network weights. This makes it possible to preserve the tree topologies when training the ForestNet. The optimized ForestNet can thus be made efficient again by mapping back to the fast and memory-compact RF version of itself as done in \cite{Richmond2015}. With this ability, we explore variants of the ForestNets where we allow different parts of the network to be optimized, in some cases retaining the ability to map-back and in other cases not. The second motivation for exploring this mapping is that it allows for the construction of an \textit{ensemble} of NNs. This ensemble is fully differentiable making it amenable to traditional NN training paradigms, unlike traditional RFs. Further, we can apply a robust averaging to the ensemble's outputs. In particular, we introduce a variant of the 3D geometric median and implement it as a fully differentiable NN module. We observe that by appending this module onto the ForestNet, either post-hoc or integrated into the end-to-end training of the network, we improve on the accuracy of the scene coordinate predictions by 6.8\% and 7.8\%, respectively, over a ForestNet without averaging. In addition to this, we compare a ForestNet to a deep Convolutional NN (CNN) trained specifically for dense scene coordinate regression. The CNN achieves better scene coordinate predictions but under-performs the RFs and ForestNets in terms of final 6D camera pose accuracy. Importantly, this deep CNN cannot be mapped to a RF, and hence cannot enjoy their fast test-time speeds and lightweight memory requirements. All these results  shed light on the best approaches to use for other continuous regression tasks.

Based on the above, our key contributions are as follows:
\begin{enumerate}
\item A new NN architecture which we term a ForestNet for dense scene coordinate regression. The ForestNet architecture is derived and initialized from a RF. We observe that for the camera localization task, the best performing ForestNet is test-time efficient with a low memory cost since it can be mapped back to a RF.
\item A fully-differentiable 3D geometric median filter implemented as a NN module for robust averaging. We append this module to a ForestNet and show that when applied post-hoc or in end-to-end training of the ForestNet, the scene coordinate accuracy improves.
\item An engineered solution to reduce the memory requirements of the ForestNets by a factor of at least four, allowing full training and testing on a single GPU.
\item An improvement over the state-of-the-art for camera localization on the 7-Scenes dataset \cite{Shotton2013} using our best method, an efficient ForestNet with robust averaging.
\end{enumerate}

\section{RELATED WORK}\label{sec:relatedwork}

\subsection{Camera (re)localization}
Camera localization has traditionally been formalized as a descriptor matching problem, addressed either using whole image-based approaches with keyframe/keypose pairs, or using sparse feature-based approaches with keypoints. In keyframe-based methods \cite{Klein2008, Gee2012,Glocker2015}, a descriptor is computed for a (whole) query image and is compared to the descriptors of a set of saved images (keyframes), each keyframe with an associated ground-truth camera pose (keypose). Keypoint-based methods instead extract interest points (keypoints) and attach a feature descriptor and 3D location to each. The camera pose of a query image is then estimated using a sparse set of keypoint 2D-to-3D correspondences \cite{Irschara2009, Lowe2004, Sattler2011, Lepetit2006}.  Despite their successes (e.g. in visual SLAM \cite{Davison2007, Se2005, Eade2008, Williams2011}), a central challenge in both approaches is the (often online) selection of keyframes and keypoints such that they provide good spatial coverage of the scene.

Departing from a model-based approach, \cite{Shotton2013} propose Scene Coordinate Regression (SCoRe) forests. They learn dense 2D pixel to 3D scene coordinate correspondences (relative to a scene-specific reference frame) using a dataset of RGB-D images and ground truth camera poses. With a sparse set of these correspondences, camera pose hypotheses are generated and a RANSAC-based optimization and refinement produces the final camera pose. A limitation of their formulation, however, is the inevitable many-to-one mapping of correspondences, making it poor at resolving scene ambiguities. \citet{Valentin2015} tackle this problem by replacing the single mode in the forest leafs with multi-modal mixtures of Gaussians, and then, at test time, using the support and spread of each mode to aid the follow-up RANSAC optimizer. Along the same vein of exploiting uncertainty for performance gains, \citet{Brachmann2016} introduce a stacked (auto-context) classification-regression version of the forest of \cite{Valentin2015}, which they use to achieve state-of-the-art results in object pose estimation and results on-par with \cite{Valentin2015} in camera localization. Unlike \cite{Valentin2015}, their results also include a case for RGB-only.

Inspired by successes in deep learning for a multitude of tasks, \cite{Kendall2015, Kendall2016} were first to propose the use of a deep CNN as an end-to-end 6D camera pose regressor using RGB and RGB-D images as input. While they achieve moderate results in large-scale outdoor scenes, their performance in small, indoor scenes is one order of magnitude worse in pose translation error than \cite{Brachmann2016} and \cite{Valentin2015}. This suggests that the intermediate step of predicting 2D-to-3D point correspondences is important for good camera pose estimation.  
\subsection{Random Forests as Neural Networks}
\citet{Sethi1990}, \citet{Welbl2014} propose a class of two-hidden-layer NNs whose architecture and initialization can be derived from a trained decision tree. The motivation behind this mapping is that it transforms a RF into a differentiable, end-to-end learnable structure which can then be further refined (i.e. fine-tuned) from a data-driven starting point when only a small amount of training data is available. \citet{Richmond2015} extend this mapping to a stacked auto-context RF, and use the equivalent deep and sparsely connected CNN to achieve improved results over the original RF stacks on two small-scale segmentation tasks. They additionally propose an approximate reverse mapping, from CNN to RF, such that the optimized network parameters can be used to update the original RF (its topology fixed) for faster test-time evaluation. Their optimized RF outperforms the original RF on the segmentation tasks, but not the mapped CNN. Our work builds on this and employs the forward mapping on a non-stacked RF trained on 2D-to-3D scene coordinates.

\subsection{Globally differentiable Random Forests}
RFs are capable of handling high-dimensional data and multi-class problems, are fast to train and test, and do not require large amounts of training data. RFs, however, are trained in a greedy fashion, with each split node optimizing some local splitting criterion conditioned on the samples that it receives. Additionally, each tree is trained independently and hence there is no principled minimization of a global loss objective across the whole forest. To address the limitations of greedy forest construction, inspired by boosted methods, \cite{SchulterADF2013} grow a forest breadth-wise, layer by layer, with the current layer's split node parameters optimized based on an across-forest loss function. \citet{Ren2015} instead propose a global post-hoc refinement and pruning of a greedily-trained RF using all training samples. They re-learn all leaf parameters and generate more compact RFs with greater generalisation capacity. \citet{Norouzi2015} look at the optimization of a single tree and present a convex-concave upper bound which they use as a proxy for a global loss function. They optimize a tree's (oblique) split functions jointly with its leaf parameters using stochastic gradient descent, imposing regularization constraints such that the hard binary decisions of the split nodes and the 1-of-$l$ encoding of leaf membership, where $l$ is the number of leafs, is preserved. Fuzzy or soft decision trees \cite{Suarez1999} relax this constraint and model split nodes as sigmoidal functions: a training sample is routed with some probability $p$ to its left child, and $1-p$ to its right child. This allows samples to be assigned partial membership to all leaf nodes rather than full membership to a single one. Much like the tree-to-NN mapping of \cite{Sethi1990}, this fuzzification allows for the otherwise local and greedy tree construction to be reframed as a (tree-wide) differentiable optimization problem. More recently, \cite{Kontschieder2015} unify the feature learning capability of deep CNNs with the ensemble nature of RFs. Similar to \cite{Suarez1999} in their use of sigmoidal split node functions, they propose a globally differentiable RF by appending a RF to a deep CNN such that the activation of each CNN output node drives the soft decision function of a RF split node. In our work, using the mapping of \cite{Sethi1990}, we explore all variants of this: learning leaf parameters only, learning leaf and split node parameters, and finally relaxing the tree constraint and allowing for leaf, split and the tree topologies to be learned.

\section{Random Forests to ForestNets} \label{subsec:RF2CNN}
In this section, we present the scene coordinate regression forests of \cite{Shotton2013} and show how they can be applied to the tree-to-NN mapping of \cite{Sethi1990, Welbl2014} for multi-dimensional regression.
\subsection{Scene coordinate regression}
For each 2D pixel in an image $\mathbf{p} \in \mathbb{R}^2$, there exists a 3D scene coordinate label $\mathbf{m} \in \mathbb{R}^3$ (see Fig.~\ref{fig:coolimage}). We can compute $\mathbf{m}$ as $\mathbf{m} = \mathit{H} \mathbf{x}$, where \textit{H} is the 6D camera pose of the image and $\mathbf{x}$ is the 3D camera coordinate obtained by back-projecting pixel $\mathbf{p}$ using its depth value. A RF is greedily trained with a set of $(f(\mathbf{p}), \mathbf{m})$ pairs, where $f(\cdot): \mathbb{R}^2 \rightarrow \mathbb{R}^D$ is the $D$-dimensional feature vector of a pixel (here $D=1000$). In training, each split node is assigned $(\boldsymbol {\theta}_n, \tau_n)$ from a pool of $N$ candidates, where $\boldsymbol{\theta}_n$ are the feature parameters (see below), $\tau_n$ is a scalar threshold and $n \in \{1,.., N\}$. Each leaf node $l$ then stores some empirical 3D distribution $P_{l}(\mathbf{m})$ of the samples that have arrived at that leaf. In \cite{Shotton2013}, each leaf distribution is represented as a set of modes found using mean-shift. \citet{Brachmann2016} and \citet{Valentin2015} instead fit 3D mixtures of Gaussians to each leaf.

More formally, the decision function at split node $s$ takes the form:
\begin{equation}
df(\mathbf{p}; \boldsymbol{\theta}_n, \tau_n) = \Big[ f_{\boldsymbol{\theta}_n} (\mathbf{p}) \geq \tau_n \Big]
\end{equation}
where $[ \cdot ]$ evaluates the boolean condition and $f_{\boldsymbol{\theta}_n}(\cdot)$ is the feature response function. If $df(\mathbf{p}; \boldsymbol{\theta}_n, \tau_n) = 0$, the sample is routed to the left child, and if $df(\mathbf{p}; \boldsymbol{\theta}_n, \tau_n) = 1$ then it is routed to the right child. In general, $f_{\boldsymbol{\theta}_n}(\cdot)$ can be a function of multiple features, however, here we opt for simple axis-aligned splits: each split node selects a single feature $d \in \{1,\cdots,D\}$ and compares it to its threshold $\tau_n$. In our case, features are simple differences of RGB pixel values:
\begin{equation}
f_{\boldsymbol{\theta}_n}(\mathbf{p}) = \mathcal{I}(\mathbf{p} +\boldsymbol{\delta}_1, c_1) - \mathcal{I}(\mathbf{p} + \boldsymbol{\delta}_2, c_2)
\end{equation}
where $\mathcal{I}$ is the image intensity value of pixel $\mathbf{p}$ offset by $\boldsymbol{\delta}_i = (x_i, y_i)$ and indexed by channel $c_i \in \{R, G, B\}$. Thus the feature parameters are $\boldsymbol{\theta}_n = \{\boldsymbol{\delta}_1, \boldsymbol{\delta}_2, c_1, c_2\}$. \\

\subsection{RF to NN mapping}
\begin{figure}[h]
\centering
\vskip 2mm
\includegraphics[scale=0.25]{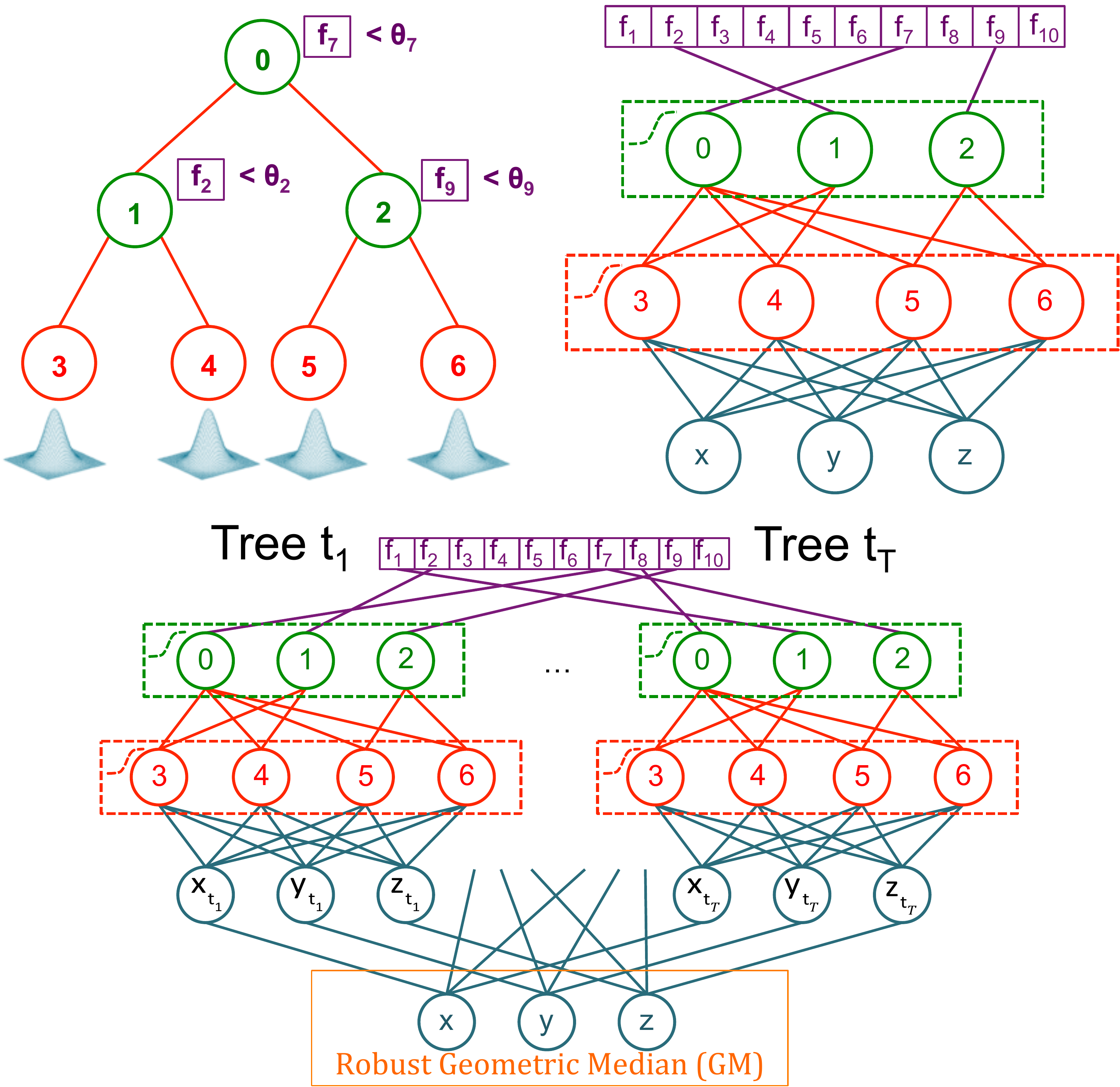}
\caption{\textbf{Forward mapping of a RF to a NN.} (Top) A single tree is mapped to a two-hidden-layer NN. At test time, each split node performs a single feature selection and a binary threshold test before routing a sample onwards. This is mimicked in the NN through the activation of specific connections. (Bottom) The single tree-to-NN mapping is replicated for each of the $T$ trees in a forest. Note: bias nodes are not shown here for readability. Purple indicates feature vector. Green indicates split layer. Red indicates leaf layer. Blue indicates output scene coordinate layer. Orange indicates robust geometric median applied to multiple NN outputs. Dotted lines indicate activation functions ($L_1: tanh(\cdot)$ and $L_2:sigm(\cdot)$ or $softmax(\cdot)$). Best viewed in color.}
\label{fig:basicRF2NN}
\end{figure}

As highlighted in \cite{Sethi1990}, the mapping from RF to NN is useful in two ways:
\begin{itemize}
\item It defines a specific NN architecture in terms of the number, structure and connections of the nodes
\item It uses the learned trees' structures and leaf distributions to initialize all network parameters.
\end{itemize}
We wish to take a trained RF for scene coordinate regression and construct an ensemble of NNs. The following describes the mapping of a single tree to a single two-hidden-layer NN (Fig.~\ref{fig:basicRF2NN})

Let a tree consist of split nodes $\tilde{s}_j \in \mathcal{\tilde{S}}$ and leaf nodes $\tilde{l}_k \in \mathcal{\tilde{L}}$. The corresponding NN will have its first layer evaluating the decision functions $df(\cdot)$ of all of the split nodes in the tree simultaneously, and its second layer encoding the leaf membership of each sample (that is, to which leaf each sample is routed).

\textbf{\textit{Split layer}}.
The split nodes $\tilde{s}_j$ of the tree form the first network layer, $L_1$, consisting of equivalent split neurons $s_j$ whereby a tree with $J$ split nodes will have $J$ neurons in $L_1$. As input to the network, each pixel $\mathbf{p}$ provides its feature vector $f(\mathbf{p}) \in \mathbb{R}^D$. Based on each split node's associated $(\boldsymbol{\theta}_n, \tau_n)$, its corresponding split neuron $s_j$ selects the feature with index $d \in \{1,...,D\}$ such that its activation $a(s_j)$ behaves as follows:
\begin{equation}
a(s_j) =
\begin{cases}
-1, & \text{if}\  f_{\boldsymbol{\theta}_n}(\mathbf{p}) < \tau_n\\
1, & \text{if}\ f_{\boldsymbol{\theta}_n}(\mathbf{p}) \geq \tau_n
\end{cases}
\end{equation}

corresponding to routing pixel $\mathbf{p}$ to the left or right child, respectively. To implement this, a sparse connection matrix is defined between the network input and $L_1$ such that the $dth$ channel of the input vector is connected to neuron $s_j$ with weight $w_{d, s_j} =  c_{01}$, where $c_{01}$ is some constant value. All other incoming weights to $s_j$ are zero (purple to green connections in Fig.~\ref{fig:basicRF2NN}). The bias of neuron $s_j$ is $b_{s_j} = - c_{01} * \tau_n$. Implemented as a simple linear layer, each neuron $s_j$ in $L_1$ thus computes:
\begin{equation}
c_{01} * f_{\boldsymbol{\theta}_n}(\mathbf{p}) - c_{01} * \tau_n
\end{equation}
and this result is assigned to $a(s_j)$ where $a(\cdot) = tanh(\cdot)$. Layer $L_1$'s activation pattern thus encodes the to-left-child or to-right-child evaluation of each split node in the tree with one forward pass through $L_1$.

The hardness of the tree's decision functions are controlled by the hyperparameter $c_{01}$: a high value of $c_{01}$ ensures that $a(s_j)$ approaches $-1$ very closely if $\tilde{s}_j$ routes a sample left, and $a(s_j)$ approaches $+1$ very closely if $\tilde{s}_j$ routes a sample right. A lower value of $c_{01}$ allows for samples to be routed partially left and right. 

\textbf{\textit{Leaf layer}}.
$L_2$, the second layer of the network, is constructed from all the leaf nodes ${\tilde{l}_k}$ in the tree, such that a tree with $K$ leaves will have $K$ leaf neurons in $L_2$. This layer must interpret the activation pattern of $L_1$ such that $L_2$ encodes the leaf to which each sample is directed.

To implement this, a sparse set of connections is constructed between $L_1$ and $L_2$ (green to red connections in Fig.~\ref{fig:basicRF2NN}). A connection exists between a split neuron $s_j$ and a leaf neuron $l_k$ if $s_j$ is on the path to $l_k$ in the tree. This means that a leaf node at depth $r$ in the tree, corresponds to a leaf neuron with $r$ incoming connections in the NN. 

The weights of these connections are such that if $\tilde{l}_k$ is in the left sub-tree of $\tilde{s}_j$ then $w_{s_j, l_k} = + c_{12}$, otherwise $w_{s_j, l_k} = - c_{12}$, where $c_{12}$ is again some constant. If $\tilde{s}_j$ is not on the path to $\tilde{l}_k$, then no connection exists (or equivalently, $w_{s_j, l_k} = 0$). With this formulation, the active leaf has incoming weights that are sign-matched to the activations on $L_1$ of the split nodes on its path. This makes the leaf neuron maximally activated. To distinguish leafs at different depths, the bias of a leaf neuron is such that $b_{l_k} = - c_{12} * (|P(\tilde{l}_k)| - 1)$ where $P(\tilde{l}_k)$ is the path length to $\tilde{l}_k$. This sets the value of the active leaf neuron to $+c_{12}$, and the values of all other leafs to $\leq -c_{12}$. The activation function $a(\cdot) = sigm(\cdot)$ is then applied to $L_2$ resulting in the $L_2$ activation pattern behaving like a binary switch, where the active leaf neuron $a(l_k)$ is $+1$, and all others are set to zero. This behaviour mimics a hard tree and can be enforced by assigning a high value to $c_{12}$. By relaxing $c_{12}$ (and $c_{01}$) and replacing $a(\cdot)$ with a \textit{softmax} activation, we can relax tree hardness.

\textbf{\textit{Scene coordinate layer}}.
Each leaf $\tilde{l}_k$ in the RFs of \cite{Brachmann2016} contains a mixture of Gaussians over 3D scene coordinates $\mathbf{m}$. Let the mode with the highest support in leaf $\tilde{l}_k$, denoted $\mathbf{m}_k \in \mathbb{R}^3$, become the 3D scene coordinate representing that leaf. At test time, a sample reaching leaf $\tilde{l}_k$ will take on label $\mathbf{m}_k$. To implement this, the network output consists of three nodes, $\mathbf{q} \in \mathbb{R}^3$, which are fully connected to all leaf neurons in $L_2$ (red to blue connections in Fig.~\ref{fig:basicRF2NN}). The weights of the three connections between leaf neuron $l_k$ and the three output nodes $\mathbf{q}$ correspond to the $x, y$ and $z$ entries of $\mathbf{m}_k$. This can be denoted as $w_{l_k, \mathbf{q}} = \mathbf{m}_k$. The output bias is $b_{\mathbf{q}} = \mathbf{0}$.  With this formulation, for a given sample, a series of zeros and a single $+1$ on the active leaf in $L_2$ result in the $\mathbf{m}_k$ of that active leaf appearing on the output nodes $\mathbf{q}$.

\textbf{\textit{Trees to ForestNets}}.
This mapping naturally extends itself to a RF with multiple trees where for a forest of $T$ trees, an ensemble of $T$ NNs can be constructed, collectively referred to as a ForestNet (Fig.~\ref{fig:basicRF2NN}). At test time, the $T$ predictions from a RF or its equivalent ForestNet can be used independently or combined to produce a single 3D prediction.

\textbf{\textit{Feature learning}}.
In recent years, it has become evident that the power of neural networks lies in their representation learning capability. Our framework is amenable to this by prepending feature learning layers to a ForestNet and by allowing linear combinations of features (purple weights) to be learned for each split node. We, however, leave this to future work, with the focus here being on the usefulness of the tree-to-NN mapping.

\section{Robust Geometric Median Averaging} \label{subsec:geomavg}

A ForestNet of $T$ NNs produces $T$ scene coordinate predictions $\mathbf{q}_{1}, \mathbf{q}_{2}, \dots ,\mathbf{q}_{T}$ which can be robustly averaged to produce a single scene coordinate, denoted $\mathbf{\tilde{q}} \in \mathbb{R}^3$. The robust average we use is
a variant of the geometric median\footnote{An extension of the median to
higher dimensions. It is defined as the point minimizing the sum of
Euclidean distances to all points in a set of discrete sample points.}.
For simplicity, we refer to it as simply the geometric median.

We start by calculating the mean of the
original predictions. With this as initialization, we continue
with a fixed number of steps of an iteratively re-weighted least squares
algorithm.
Each iteration calculates a weighted average:
\begin{equation}
\mathbf{\tilde{q}}^{t+1}= \frac{\sum_{i=1}^{T} w_i^t \mathbf{q}_i }
{\sum_{i=1}^{T} w_i^t }
\end{equation}
of the original predictions. For the first 10 iterations we use the weights:
\begin{equation}
w^t_i=\frac{1}{\| \mathbf{\tilde{q}}^{t} -  \ \mathbf{q}_i \|_2}
\end{equation}
This is equivalent to the Weiszfeld algorithm \cite{Weiszfeld2009}
which approximates the geometric median.  While this result can
be used directly as a robust average, we find that it is beneficial to
apply a further 10 iterations using weights with the form:
\begin{equation}
w^t_i=\exp(- \frac{\| \mathbf{\tilde{q}}^{t} -  \mathbf{q}_i \|_2^2}
{2\sigma^2} )
\end{equation}
corresponding to a mean-shift algorithm with a Gaussian kernel of
standard deviation $\sigma$ (here $\sigma=2.5cm$). The iterations  converge to a locally
dominant mode. The result of the last iteration is our final robust
average $\mathbf{\tilde{q}}$.

Note, that since each of the iterations above is simply a weighted average, the entire process is fully differentiable which allows us to
implement the robust average as a multi-layer module (which we call
GM) which has no learnable parameters. We investigate the use of our robust averaging in two
different settings:
\begin{enumerate}
\item The GM module is appended to the ForestNet and the full network
trained end-to-end. We call this eGM (for end-to-end).
\item Each ForestNet tree is trained independently and the GM module is appended post-hoc at test-time. We
call this pGM (for post-hoc).
\end{enumerate}
We apply these two types of robust averaging to all ForestNet variants,
as well as to a modified RF from \cite{Brachmann2016}. Note, however,
with an RF averaging can only be done post-hoc (see RF2-pGM in Section~\ref{subsec:methods}).

\section{Network Splitting}\label{subsec:NNsplitting}

Although the RF-to-NN mapping enforces sparse connectivity between layers $L_1$ and $L_2$, sparse operation on GPU is not always efficient. For this reason, $L_1$ and $L_2$ are implemented as fully connected layers with the inactive connections set and held at zero. The state-of-the-art results of \cite{Brachmann2016} use RFs of $3-5$ trees with depths between $15$ and $16$. This corresponds to $\mathtt{\sim}37 000$ split and $\mathtt{\sim}37 000$ leaf nodes per tree. This constructs a network on the order of $1.4$ billion parameters per tree. Practically, with floating-point operations, this requires $5.6GB$ of memory per tree, rapidly making the full ForestNet difficult to fit on a single GPU.
\begin{figure}[thpb]
\centering
\vskip 4mm
\includegraphics[scale=0.27]{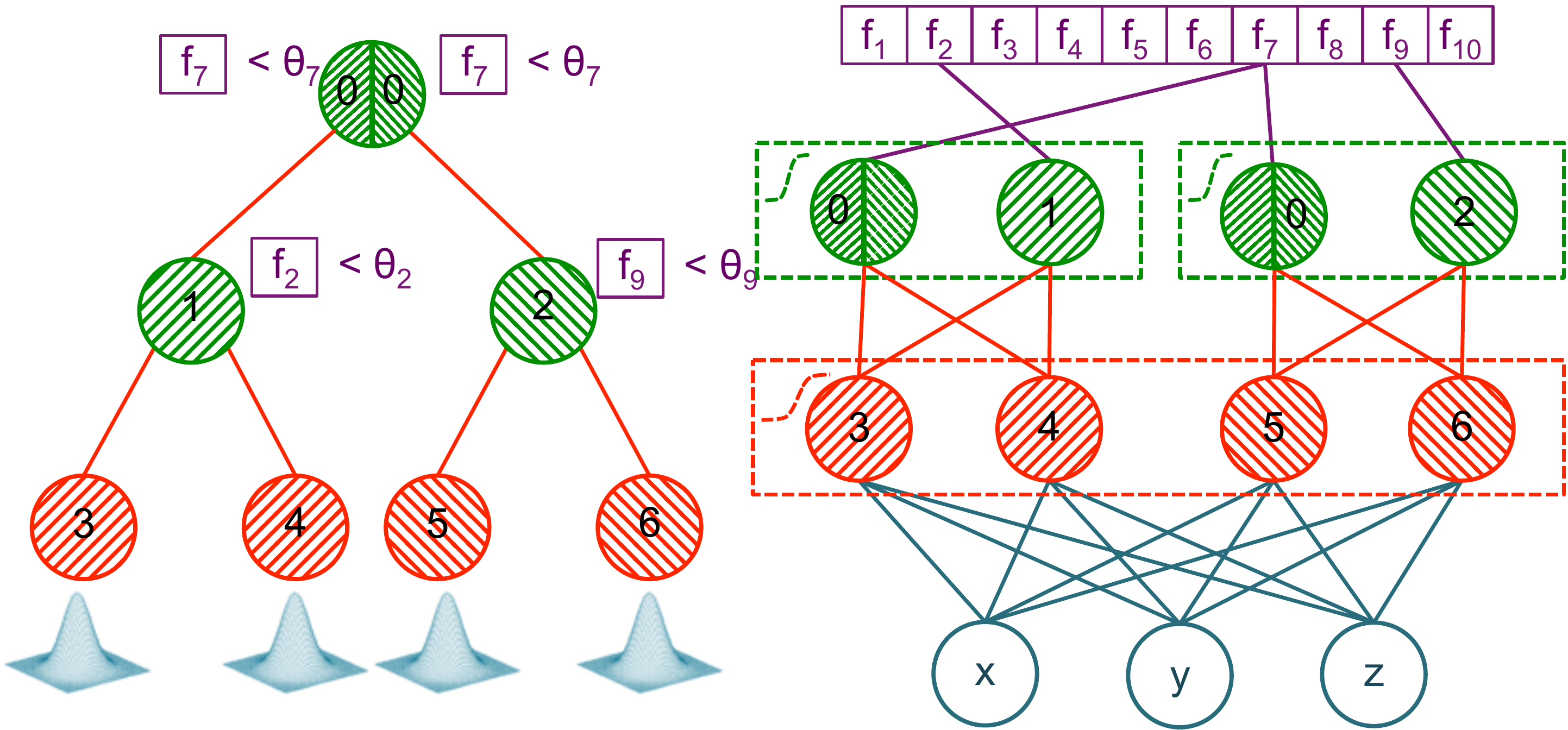}
\caption{\textbf{Network splitting for efficient GPU training/testing}. Trained trees are decomposed into sub-trees and mapped to NNs with weights shared across the common root nodes. (Left) A tree is split into 2 sub-trees ($r=3, \tilde{r}=2$). (Right) The construction of its equivalent NN. The weight of the root node (0) is shared between the two sub-NNs. Hatching direction indicates sub-tree membership. Best viewed in color.}
\label{fig:splitnetworks}
\end{figure}
To tackle this, a network splitting strategy is implemented whereby each trained tree is decomposed into a set of sub-trees, each mapped to a sub-NN with the weights of the sub-trees' root nodes shared across the sub-NNs (see Fig.~\ref{fig:splitnetworks}). If we let the original tree depth be $r$, and the depth of the sub-trees be $\tilde{r}$, such that $0 < \tilde{r} \leq r$, then the total number of parameters in the ForestNet can be reduced by a factor of:
\begin{equation}
\frac{2^r - 1}{2^{\tilde{r}} - 1 + r-\tilde{r}}
\end{equation}
Using this splitting strategy, we are able to reduce each full ForestNet from
$\mathtt{\sim}28GB$ to $\mathtt{\sim}7GB$, allowing for effective training and testing on a single GPU.\\

\section{EXPERIMENTS}
\label{sec:experiments}

\subsection{Datasets}
We use the 7-Scenes dataset \cite{Shotton2013} which contains RGB-D images captured with a handheld Kinect camera ($640\times480$ resolution) and associated ground-truth camera pose calculated using the Kinect Fusion implementation of \cite{Newcombe2011}. RGB and depth images are manually registered. Depth images are used to pre-compute the ground truth scene coordinates per frame, however, only RGB images are used at test-time.

\subsection{Hardware}
We train and test all NNs on \textit{NVIDIA GeForce GTX TitanX} and all RFs on \textit{Intel}{\textregistered}Core{\texttrademark} i7-3770K CPU $@ 3.5GHz$.

\subsection{Performance metrics}
We measure performance on (1) the accuracy of the scene coordinate predictions, and (2) the final camera pose accuracy. We quantify (1) in terms of the number of scene coordinate predictions considered to be inliers and the mean Euclidean distance of these inliers from their ground truth scene coordinate labels. A prediction is counted as an inlier if its Euclidean distance to its ground truth label is $< 10cm$. We quantify (2) in the same way as \cite{Shotton2013, Brachmann2016, Valentin2015} with a $5cm \; \& \; 5^\circ$ criterion: the camera pose is taken to be correct if it is within $5cm$ translation and $5^{\circ}$ rotation from the ground truth camera pose. We report the percentage of correct camera poses as well as the average median camera error (we calculate the median per scene and average across scenes).
\subsection{Methods}
\label{subsec:methods}
We define the following baselines:
\begin{itemize}
\item The current state-of-the-art in camera localization \cite{Brachmann2016} which we refer to as \textbf{RF1}.
\item A modified version of RF1 which we call \textbf{RF2}. RF2 has 5 rather than 3 trees, no auto-context and its leafs hold only the single highest-supported mode. RF2 is used to map to the ForestNets.
\item A deep CNN trained to predict 2D-to-3D scene coordinates in a patch-like manner. This CNN's architecture is based on AlexNet \cite{Krizhevsky2012} but is trained from scratch. Instead of using the standard two channel architecture, we use a single channel with double capacity (96 filters in the first \textit{conv} layer and 4096 neurons in the fully-connected layers). We also replace the 1000 neuron output layer with 3 neurons to predict a single 3D scene coordinate. We refer to this as \textbf{D-NET} (for deep).
\end{itemize} 

We also explore the following three variants of ForestNets:
\begin{itemize}
\item \textbf{ForestNet Leafs (fNET-L)} in which only the leaf weights (blue) are learned. fNET-L preserves the tree topologies and can be mapped back to a fast and memory-efficient RF with optimized leaf modes.
\item \textbf{ForestNet Leafs \& Splits (fNET-LS)} in which the leaf parameters and the split node thresholds (i.e. the biases between the network input and $L_1$) are learned. fNET-LS also preserves tree topology, however, since the split thresholds can be adjusted, it is likely that multiple leaf nodes are activated. Because of this, mapping back to an RF is possible but only an approximation \cite{Richmond2015}.
\item \textbf{ForestNet Leafs, Splits \& Topology (fNET-LST)} in which the leaf parameters, split node thresholds and the tree topologies (red weights) are learned. fNET-LST cannot be mapped back to a RF since the tree structures are not preserved.
\end{itemize}

For all of our methods, during the second stage of RANSAC-based pose optimization, we increase the number of pose hypotheses drawn from 256 (as used in \cite{Brachmann2016}) to 1280 to increase the chance of finding a good solution.
\def\arraystretch{1}
\begin{table*}[thpb]
\begin{center}
\centering
\vskip 2mm
\caption{\scriptsize{\linespread{0.5}Coordinates: mean inlier count (brackets show mean Euclidean distance error of inliers). Camera pose: median translation and rotation error, averaged over scenes (brackets show percentage of poses meeting $5cm \; \& \; 5^\circ$ criterion)}}
\label{tab:quants}
\begin{tabular}{|M{0.15\columnwidth}|M{0.15\columnwidth}||M{0.22\columnwidth}|M{0.22\columnwidth}|M{0.22\columnwidth}|M{0.22\columnwidth}|M{0.23\columnwidth}|M{0.22\columnwidth}|M{0.22\columnwidth}|}
\cline{1-9}
\multicolumn{1}{|c|}{} &        & RF1 \cite{Brachmann2016} & RF2 & fNET-L & fNET-LS & fNET-LST & D-NET & PoseNet \cite{Kendall2015}\\ \cline{1-9}
Overview & Description & state-of-the-art RF \cite{Brachmann2016} & RF of \cite{Brachmann2016} with adapted parameters  &fNET which learns leafs. Can be mapped-back to RF & fNET which learns leafs \& splits. Can be approximately mapped back to RF & fNET which learns leafs, splits \& tree topology. Cannot be mapped back to RF & NN for dense scene coordinate regression & NN for direct 6D camera pose regression \\ \cline{2-9} 
\multicolumn{1}{|c|}{}  & Speed  &   fast  &   fast  &    fast   &        slow* &      slow    &  slow & slow       \\ \cline{2-9} 
\multicolumn{1}{|c|}{} & Memory & low     &  low   &  low     &       high*  &  high         &   high & high      \\ \hline \hline
Accuracy w.r.t  & noGM & 21.6\% (5.2cm) & 18.3\% (4.9cm)  &  16.8\% (4.9cm) &  13.4\% (5.6cm) & 17.3\% (5.6cm) &  45.3\% (4.7cm) & n/a \\
Scene & pGM & n/a &  26.1\% (4.7cm) &  23.6\% (4.7cm) &   17.9\% (5.4cm)  & 21.1\% (5.5cm) & n/a & n/a  \\
Coordinates& eGM & n/a & n/a & 24.6\% (4.8cm)  &  21.0\% (4.9cm) &  12.9\% (5.5cm) & n/a & n/a  \\
\hline
Accuracy w.r.t.  & noGM & $6.1cm\;2.7^\circ$ (55.2\%) & $4.2cm\;2.1^\circ$ (62.6\%) & $4.6cm\;2.2^\circ$ (61.3\%) & $7.6cm\;2.9^\circ$ (36.7\%) & $6.2cm\;2.9^\circ$ (50.5\%) & $4.6cm\;2.1^\circ$ (57.7\%) & $44.0cm\;10.4^\circ$ \\
Camera  & pGM & n/a & $3.8cm\;1.9^\circ$ (64.5\%) & $3.9cm\;1.9^\circ$ (63.9\%) & $6.7cm\;2.5^\circ$ (39.1\%) & $5.3cm\;2.6^\circ$ (52.8\%) & n/a & n/a  \\
Pose & eGM & n/a  &  n/a & $4.4cm\;2.1^\circ$ (62.0\%) & $4.5cm\;2.2^\circ$ (60.4\%) & $12.2cm\;7.0^\circ$ (39.6\%)& n/a & n/a \\
\cline{1-9}
\cline{1-9}
\end{tabular}
\vskip 0.2cm
{*Approximate map back to fast, memory-efficient RF is possible, but with potential performance loss}
\end{center}
\vskip -5mm
\end{table*}

\subsection{Training}
\label{subsec:training}
We train ForestNets with a loss on the Euclidean distance between the prediction $\mathbf{q}$  (or robust prediction $\mathbf{\tilde{q}}$) and ground truth scene coordinate $\mathbf{m}$:
\begin{equation}
\mathcal{L} = \sum_{i=1}^N \| \mathbf{q} - \mathbf{m} \|_2
\end{equation}
where $N$ is the number of training samples. Training is done using standard backpropagation via stochastic gradient descent, with batches of size 20 and a learning rate of 0.001.  

\section{Results}
Table~\ref{tab:quants} shows our quantitative results. We summarize our key observations as follows:  
\begin{itemize}
\item Of the fast methods, RF2-pGM (RF with post-hoc geometric median) is the best performing. It supersedes the current state-of-the-art of \cite{Brachmann2016} by 7.8\% in inlier count and 9.3\% in proportion of correct final camera poses.

\item Of the ForestNet variants, fNET-L achieves the best inlier count of 24.6\% (with eGM) and the best camera pose error of $3.9cm \: 1.9^\circ$ (with pGM). While its inlier count is not as good as RF2-pGM, fNET-L-pGM achieves on-par performance with RF2-pGM in the final camera pose accuracy. fNET-L-pGM can thus be used as an equivalent and importantly differentiable replacement of a traditional RF. This would allow, for example, a ForestNet to be prepended to a differentiable RANSAC module (see recent work by \cite{Brachmann2016b}) thus making a complete end-to-end differentiable camera pose estimation pipeline.

\item Our robust geometric median filter offers notable performance gains. Across all RF and ForestNet methods, relative to noGM, inlier counts improve by 5.7\% with pGM and 3.7\% with eGM. The proportion of correct final camera poses improves by 2.4\% with pGM and 4.6\% with eGM. The non-deterministic nature of RANSAC obscures the true reason for the improvement of eGM over pGM in final camera pose (since it is inferior in inlier count performance), however it is clear that geometric median robust averaging offers gains over using no robust averaging at all.

\item D-NET supersedes all of the noGM methods in terms of scene coordinate performance, with a significantly higher number of inliers than both the RFs and ForestNets. This, however, does not translate to D-NET having the best camera pose accuracy. Overall, we observe that scene coordinate accuracy and final camera pose accuracy are only mildly correlated, for the reason mentioned above. 

\item The two-part pipeline with intermediate scene coordinate prediction has significant gains over direct camera pose regression. All our methods exceed the final camera pose accuracy of \cite{Kendall2015} by an order of magnitude. 
\end{itemize} 

\textbf{Speed and memory}. A 5-tree RF has a model size of $\sim$250MB and can obtain dense coordinate predictions for a frame on CPU in 100-150ms (equivalently 5-10ms on GPU). This is low memory and fast. An equivalent ForestNet has a model size of $\sim$7GB (although many parameters are fixed at zero) and it can densely process a frame in a patch-like manner in 5-7 minutes on GPU. This is high memory and slow. If the ForestNet can be mapped back to an RF (like fNET-L and fNET-LS) then it can be transformed to a fast and lightweight version of itself. A D-NET can be stored in $\sim$500MB and can densely process an image in 1-2s. Unlike a ForestNet, however, a D-NET cannot be mapped to a RF.

\textbf{Learning tree parameters}. The RFs of \cite{Brachmann2016} and \mbox{fNET-L} differ in the way that their leaf modes are optimized: with fNET-L, only the leaf weights are optimized during training. With our training loss, we optimize the sum of Euclidean distances of all samples reaching a leaf. This is equivalent to calculating the geometric median of all samples in that leaf. This is different (and seemingly inferior) to running mean-shift on the set of samples and choosing the mode with largest support, as the RFs do. Although the geometric median is robust to outliers, they still influence the prediction to some extent, which is not the case with mean-shift. This interpretation no longer holds when we train with the appended geometric median layer (eGM). In this case, the leaf predictions can adapt in a way that the resulting geometric median is accurate. In general, this improves the predictions' accuracy. We also observe that initialising the ForestNets as soft trees and allowing them to learn more parameters (fNET-LS and fNET-LST) generally negatively affects the scene coordinate and camera pose accuracy. This suggests that with too many degrees of freedom, the ForestNets cannot be properly optimized from their RF-initialized starting points.

\textbf{Scene coordinates to camera pose}. The nature of the relationship between scene coordinate accuracy and final camera pose accuracy is unclear in that better scene coordinates do not always result in better camera poses. D-NET produces a much higher number of inliers (45.3\% versus 18.3\% of RF2 and 16.8\% of fNET-L) with a better inlier error (4.7cm versus 4.9cm) yet does not achieve the best final camera pose. We attribute this to the non-deterministic and highly robust RANSAC-based optimizer.



\section{Conclusions and Future Work}
\label{sec:conslusions}

In this work we explored efficient versus non-efficient and RF- versus non-RF-derived NN architectures for camera localization. While a traditional NN architecture is superior with respect to dense scene coordinate regression, its inefficiency in terms of speed and memory makes it unattractive for mobile and robotics applications. On the other hand, our best-performing ForestNet, with a robust 3D geometric median-based average, is test-time efficient since it can be mapped back to a RF, and has improved the current state-of-the-art in camera localization on the 7-Scenes dataset \cite{Shotton2013}. Overall, however, the nature of the relationship between good scene coordinate predictions and a good final 6D camera pose is not yet clear. This motivates further research into (1) better suited loss functions for the follow-up RANSAC optimizer, and (2) the end-to-end training of a full camera pose estimation pipeline.


\addtolength{\textheight}{-9cm}   



\vspace{-2mm}
\section*{Acknowledgment}
\vspace{-2mm}
This work was supported by the EPSRC, ERC grant ERC-2012-AdG 321162-HELIOS, EPSRC grant Seebibyte EP/M013774/1, EPSRC/MURI grant EP/N019474/1 and The Skye Foundation.
\vspace{-5mm}

\bibliographystyle{IEEEtranN}
\bibliography{IEEEabrv,references}

\end{document}